\documentclass[10pt,twocolumn,letterpaper]{article}

\usepackage{cvpr}              
\definecolor{cvprblue}{rgb}{0.21,0.49,0.74}
\usepackage[pagebackref,breaklinks,colorlinks,allcolors=cvprblue]{hyperref}
\usepackage{amsmath}
\usepackage{multirow}
\usepackage{colortbl}
\usepackage{tcolorbox}
\usepackage{adjustbox}
\usepackage{booktabs}
\usepackage{algorithm}
\usepackage{algorithmic}
\usepackage{hyperref}

\definecolor{light-gray}{gray}{0.6}
\definecolor{front-color}{HTML}{F5FFFA}
\definecolor{Gray}{gray}{0.93}
\usepackage{bibentry}
\usepackage{enumitem}
\usepackage{amssymb,mathrsfs}
\usepackage{fdsymbol}
\usepackage[accsupp]{axessibility}
\usepackage[T1]{fontenc}
\usepackage{times}
\def\paperID{3700} 
\def\confName{CVPR}
\def\confYear{2026}

\title{Beyond Perceptual Shortcuts: Causal-Inspired Debiasing Optimization for Generalizable Video Reasoning in Lightweight MLLMs}
\author{Jingze Wu, Quan Zhang$^*$, Hongfei Suo, Zeqiang Cai, Hongbo Chen$^*$ \\
Sun Yat-sen University, China\\
{\tt\small \{wujz3,suohf,caizq5\}@mail2.sysu.edu.cn, \{zhangq689,chenhongbo\}@mail.sysu.edu.cn}
}

\begin{document}
\maketitle
\let\thefootnote\relax
\footnotetext{$^*$ Corresponding authors.}
\begin{abstract}
Although reinforcement learning (RL) has significantly advanced reasoning capabilities in large multimodal language models (MLLMs), its efficacy remains limited for lightweight models essential for edge deployments.
To address this issue, we leverage causal analysis and experiment to reveal the underlying phenomenon of perceptual bias, demonstrating that RL-based fine-tuning compels lightweight models to preferentially adopt perceptual shortcuts induced by data biases, rather than developing genuine reasoning abilities.
Motivated by this insight, we propose \textbf{VideoThinker}, a causal-inspired framework that cultivates robust reasoning in lightweight models through a two-stage debiasing process. First, the \textbf{Bias Aware Training} stage forges a dedicated ``bias model'' to embody these shortcut behaviors. Then, the \textbf{Causal Debiasing Policy Optimization} (CDPO) algorithm fine-tunes the primary model, employing an innovative repulsive objective to actively push it away from the bias model's flawed logic while simultaneously pulling it toward correct, generalizable solutions.
Our model, \textbf{VideoThinker-R1}, establishes a new state-of-the-art in video reasoning efficiency. For same-scale comparison, requiring \textbf{no} Supervised Fine-Tuning (SFT) and using only \textbf{1}\% of the training data for RL, it surpasses VideoRFT-3B with a \textbf{3.2}\% average gain on widely-used benchmarks and a 7\% lead on VideoMME. For cross-scale comparison, it outperforms the larger Video-UTR-7B model on multiple benchmarks, including a 2.1\% gain on MVBench and a 3.8\% gain on TempCompass. Code is available at \href{https://github.com/falonss703/VideoThinker}{https://github.com/falonss703/VideoThinker}.

\end{abstract}
\section{Introduction}
\begin{figure}[!t]
    \centering
    \includegraphics[width=0.99\linewidth]{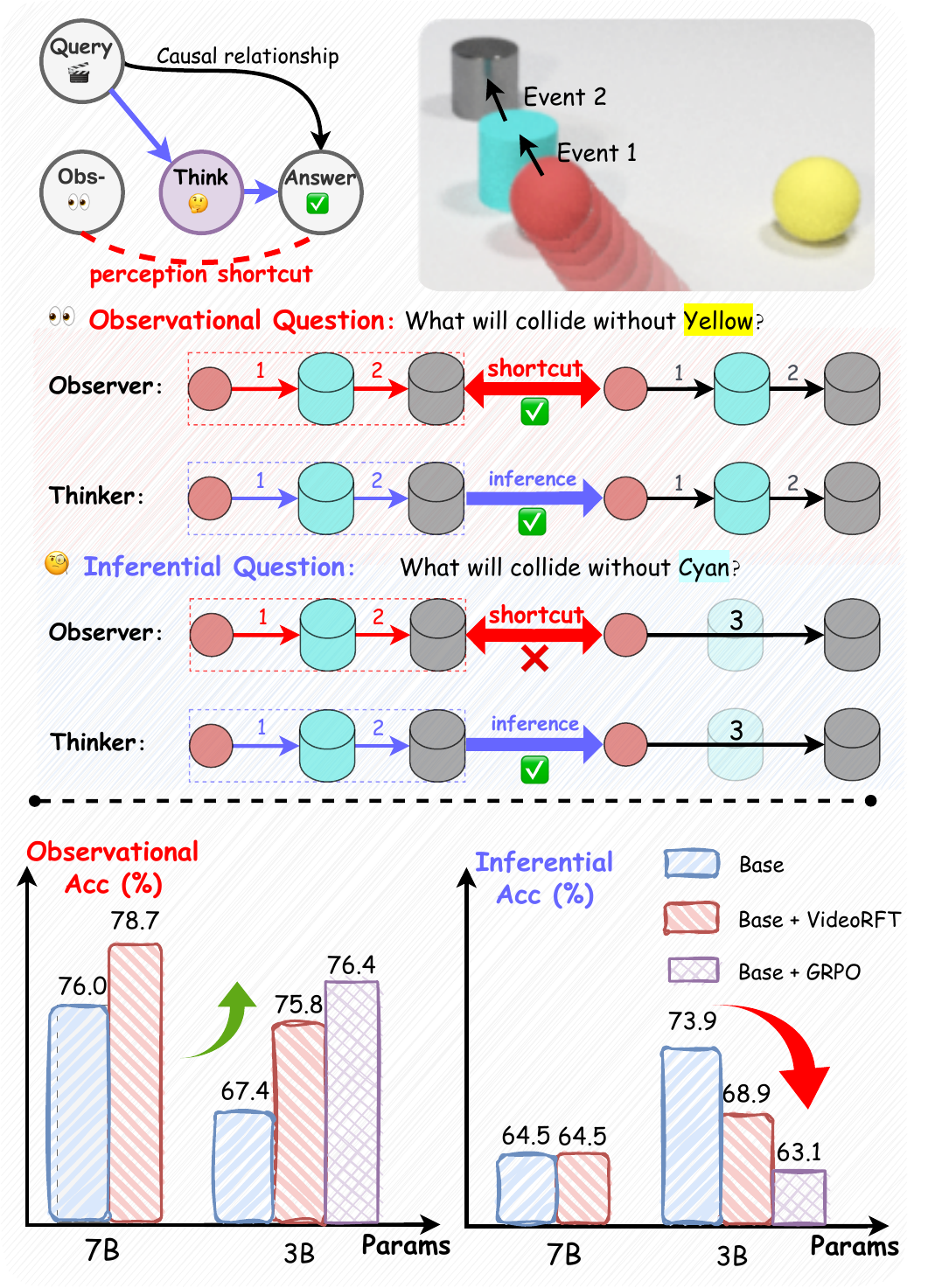}
    \caption{An illustration of observational versus inferential reasoning and the trade-off in model performance. The \textbf{top} half defines observational questions that can be answered via perceptual shortcuts and inferential questions that require true reasoning. The \textbf{bottom} shows that fine-tuning boosts accuracy on observational tasks at the expense of the lightweight model's inference.}
    \label{fig:intro}
    \vspace{-4mm}
\end{figure}

\begin{figure*}[!t]
    \centering
\includegraphics[width=1.0\linewidth]{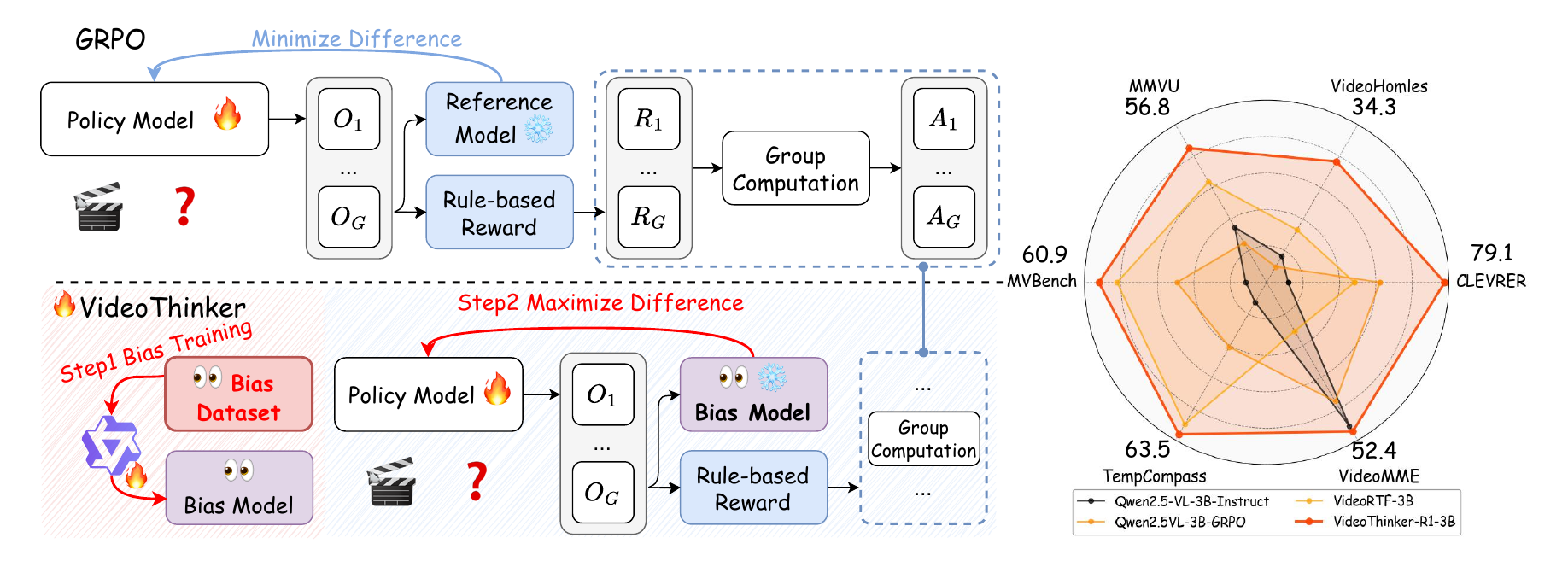}
    \caption{The difference between our framework and others. \textbf{(Left)} A conceptual comparison between our approach and a conventional baseline. (\textit{Top left}) Standard methods like GRPO constrain the policy by pulling it towards a reference model. (\textit{Bottom left}) In contrast, VideoThinker employs a ``bias model'' and a repulsive objective to push the policy \textit{away} from learned shortcuts. \textbf{(Right)} The radar chart demonstrates the empirical success of this approach, showing that our model, VideoThinker-R1, achieves state-of-the-art (SOTA) performance across multiple video reasoning benchmarks.}
    \label{fig:model_performance}
    \vspace{-4mm}
\end{figure*}

Enabling complex reasoning in lightweight Multimodal Language Models (MLLMs) is critical for their deployment in resource-constrained environments, yet it presents a formidable challenge. While fine-tuning methods based on Reinforcement Learning (RL), such as Group Relative Policy Optimization (GRPO)~\cite{shao2024deepseekmath}, have shown significant success in advancing large-scale MLLMs~\cite{xiao2025fastslow,wang2025timer, videor1, li2025videochat,dang2025reinforcing}, their effectiveness surprisingly plummets when applied to smaller models~\cite{zhang2025tinyllava, wang2025videorft}. This efficacy gap presents a critical bottleneck and raises an interesting question: Why do these proven RL techniques fail on these models?

Our investigation suggests the answer lies in a fundamental flaw within the training data. We identify a critical problem we term ``perceptual shortcuts,'' which we find are alarmingly dominant even in widely used reasoning datasets like CLEVRER~\cite{yi2019clevrer}. As illustrated in Figure~\ref{fig:intro}, our analysis reveals that many questions are merely Observational. For instance, a ``without the yellow ball'' query is non-inferential as the ball is causally irrelevant, allowing a model to succeed with a simple visual description (the shortcut). This contrasts sharply with Inferential questions, such as ``without the cyan cylinder,'' where the object acts as a causal blocker,  making true reasoning mandatory. We hypothesize that this shortcut bias is the primary culprit, and that this problem is acutely amplified in 3B models. We argue their weaker foundational capabilities compared to larger 7B models~\cite{chen2024multi,shukor2024beyond,xu2025more} make them inherently more susceptible to being misled. To verify this, we designed a diagnostic experiment. The results revealed a striking ``capability conflict.'' When a 3B base model~\cite{bai2025qwen2} with strong innate skills was fine-tuned using GRPO, its accuracy on the critical inferential tasks plummeted from 73.9\% to 63.1\%. This confirms our hypothesis: the RL process, misguided by the bias, actively forced the vulnerable 3B model to unlearn its reasoning in favor of the shortcut.

To resolve this critical fine-tuning dilemma and unlock the cognitive potential of lightweight MLLMs, our approach begins with a formal causal analysis. We construct a structural causal model that models the fine-tuning process, which explicitly identifies a spurious ``shortcut'' pathway responsible for the model learning superficial correlations. Guided by this analysis, we propose the \textbf{VideoThinker} framework, a novel framework designed to ignite robust reasoning by acting as a targeted intervention to block this undesirable path. VideoThinker achieves this through two core components, as shown in Figure~\ref{fig:model_performance}. First, to operationalize the shortcut pathway, we construct a ``bias model'' trained specifically to emulate this behavior, serving as a negative exemplar of the reasoning to be avoided. Second, during fine-tuning, Causal Debiasing Policy Optimization (CDPO) employs an innovative repulsive objective. By using a negative Kullback-Leibler (KL) divergence coefficient, it transforms the regularizer into a repulsive force, actively pushing the primary model's policy away from the bias model's. This dynamic interplay, where reward attracts the model to correct answers and repulsion steers it from bias, which compels it to develop generalizable reasoning.

Extensive experiments validate the superiority of our VideoThinker framework, as embodied by the performance of our model, VideoThinker-R1 (3B). In same-scale benchmarks, VideoThinker-R1 establishes a new SOTA, achieving 60.9\% on MV-Bench and 63.5\% on TempCompass. This strong performance extends to cross-scale comparisons, where it outperforms the much larger Video-UTR-7B model by 2.1\% and 3.8\%, respectively, showcasing a remarkable leap in reasoning efficiency. The impact of CDPO's causal debiasing is even more pronounced on tasks requiring pure causal inference. On CLEVRER, VideoThinker-R1 attains 79.1\% accuracy with merely 1k training samples, gaining a 14.2\% absolute improvement over the GRPO baseline, demonstrating its ability to learn robust reasoning from limited data.

Our main contributions are summarized as follows:
\begin{itemize}

\item We are the first to identify and verify that the ``perceptual shortcut'' phenomenon is a critical bottleneck hindering the reasoning development of lightweight MLLMs. From a causal perspective, we provide a formal analysis of the backdoor path connecting the question, reasoning, and answer, pinpointing data bias as its root cause.

\item We propose the VideoThinker framework, which first trains a ``bias model'' to embody shortcuts, then employs our CDPO algorithm with a repulsive objective to steer the main model toward genuine reasoning.

\item VideoThinker-R1 establishes a new state-of-the-art for lightweight MLLMs video reasoning task. Trained with CDPO and only 1\% of data for RL, it significantly outperforms both same-scale baselines (e.g., +7\% on VideoMME vs. VideoRFT-3B) and much larger models like Video-UTR-7B.

\end{itemize}
\section{Related Works}

\subsection{Reinforcement Learning in MLLMs}
Recent advancements in enhancing the reasoning abilities of Multimodal Large Language Models (MLLMs) have heavily relied on Reinforcement Learning (RL) frameworks. Techniques like GRPO~\cite{shao2024deepseekmath}, an efficient variant of Proximal Policy Optimization (PPO), have been widely adopted to fine-tune models for complex visual reasoning tasks~\cite{fu2024video,li2025videochat,zhan2025visionr1}. However, the predominant focus of this line of work has been on refining the optimization process itself, such as the learning algorithm~\cite{dang2025reinforcing} or the reward signal~\cite{videor1,li2025videochat,wang2025videorft}, while largely overlooking a more fundamental issue: the inherent biases and spurious correlations embedded within the training datasets~\cite{torralba2011unbiased,zhang2021seeing,zhang2024separable,zhang2025debiasing}. Consequently, even with powerful optimization, models often learn to exploit spurious correlations. For instance, they learn to associate superficial features with answers rather than developing genuine causal understanding. This leaves a critical gap, as existing methods lack an explicit mechanism to counteract shortcut learning, limiting their ability to generalize and perform true causal reasoning.

\subsection{Shortcut Learning in MLLMs}
Shortcut learning is a recognized core challenge in MLLMs, primarily manifesting as intermodal bias, where models rely excessively on language priors while neglecting visual evidence \cite{si2022language, Zang_2023_CVPR, zhou2024mitigating, park2025assessing}. To resolve this intermodal conflict, causal inference has been widely used for both diagnosis \cite{niu2021counterfactual, chen-etal-2024-quantifying} and intervention \cite{su2023language,tai2023link, zhao2023causal, chen2025cross}. However, these causal debiasing strategies are predominantly designed to resolve the intermodal conflict (i.e., whether to trust text versus vision), operating on the implicit assumption that once the model attends to the correct modality, reasoning will proceed correctly. This assumption is flawed, as it overlooks a distinct failure mode: the intramodal ``perceptual shortcut.'' While this concept is well-established in traditional image recognition tasks (e.g., relying on image background without learning robust features) \cite{qin2021causal,zhang2021deep,zhang2021learning,zhang2023wavelet,zhang2024view}, our work is the first to specify and diagnose its more advanced form in MLLMs reasoning. We find models exploit ``Observational shortcuts'' from the original video, such as merely confirming two objects never touched, to ``cheat'' on an ``Inferential question'' that should require complex counterfactual reasoning. This demonstrates that existing causal solutions are insufficient, as they are unequipped to address the deeper, intramodal ``reasoning deficit.''
\section{Methodology}
\begin{figure*}[!t]
    \centering
    \includegraphics[width=0.95\linewidth]{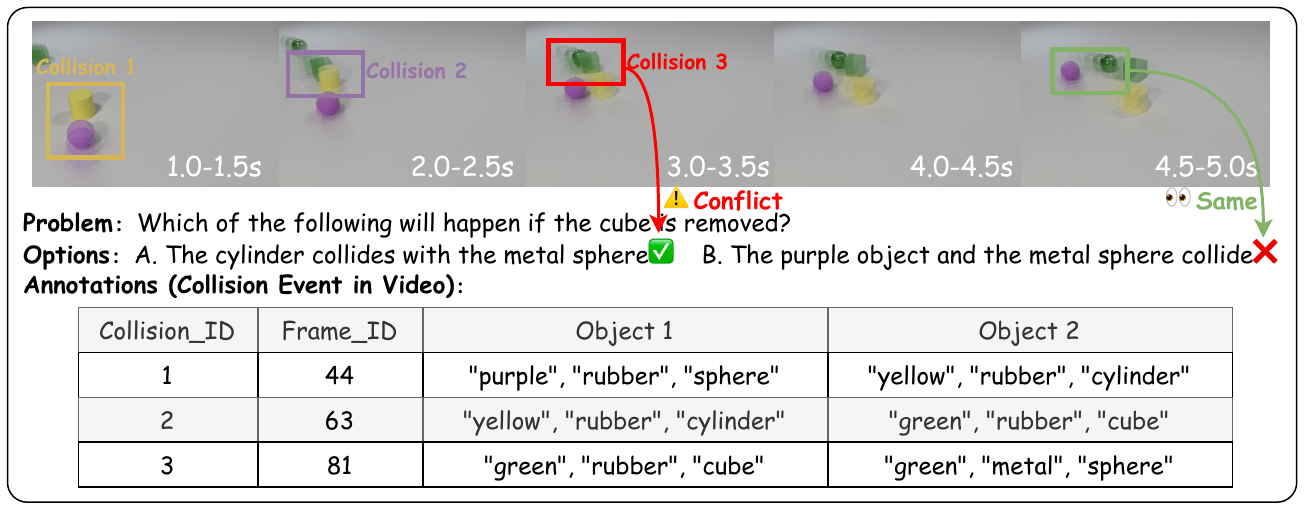}
    \caption{Illustration of Perceptual Shortcuts in Counterfactual Reasoning Tasks. The video depicts a multi-object collision scenario. Option A (Inferential) requires causal reasoning as removing the cube leads to a conflicting outcome, while Option B (Observational) allows for a ``perceptual shortcut'' by simply observing the original video.}
    \label{fig:diagnostic_chart}
    \vspace{-5mm}
\end{figure*}

\subsection{Preliminary}
\label{subsec:Preliminary}
We focus on the video reasoning task~\cite{fu2024video,hu2025video} of multi-choice question-answering (MCQA). When give a query $\mathcal{Q}$ which includes a video input, a question and the candidate answers $\mathcal{A}$, the model $f_{\theta}$ is required to predict the correct answer $a^{*}\in\mathcal{A}$ based on both the context, which can be formulated as:
\begin{equation}
    a^{*}=f_{\theta}(\mathcal{Q},\mathcal{A}).
\end{equation}
To enhance the complex reasoning abilities of MLLMs required for this task, recent works~\cite{videor1,li2023videochat,dang2025reinforcing,wang2025videorft} have moved towards RL to fine-tune large models. Notably, GRPO \cite{shao2024deepseekmath} has been successfully used to achieve state-of-the-art performance in math and coding reasoning~\cite{guo2025deepseek,yu2025dapo}. GRPO provides an efficient, criticism-free RL framework by optimizing policy-level group comparisons. It first samples a group of $G$ candidate responses $\{O_1,\dots,O_G\}$ and uses the rule-based reward model to generate scores $\{R_1, \dots, R_G\}$. These are then normalized into the relative quality:
\begin{equation}
\label{eq:ro}
    \hat{A_i}=
    \frac{R_i-\mathrm{mean}(\{R_i\}_{i=1}^G)}{\mathrm{std}(\{R_i\}_{i=1}^G)} \text{,}
\end{equation}
where $\hat{A_i}$ as the standardized advantage of the $i$-th response within the group. The policy $\pi_\theta$ is then updated by maximizing the GRPO objective function:
\begin{equation}
\begin{aligned}
\mathcal{J}_\text{GRPO}(\theta) =& \mathbb{E}_{(q,a)\sim \mathcal{D}, \{o_i\}_{i=1}^G\sim \pi_{\theta_\text{old}}(\cdot\mid q)} \\&
\Bigg[ \frac{1}{G}\sum_{i=1}^{G} \frac{1}{|o_i|}\sum_{t=1}^{|o_i|} \Bigg( 
\min \Big( r_{i,t}(\theta) \hat{A}_{i,t}, \\&
\ \text{clip} \Big( r_{i,t}(\theta), 1 - \varepsilon, 1 + \varepsilon \Big) \hat{A}_{i,t} \Big)
\\&- \beta D_{\text{KL}}(\pi_{\theta} || \pi_{\text{ref}}) 
\Bigg) \Bigg],
\label{eq:grpoloss}
\end{aligned}
\end{equation}
where
\begin{equation}
    r_{i,t}(\theta)=\frac{\pi_{\theta}(o_{i,t} \mid q, o_{i,<t})}{\pi_{\theta_{\text{old}}}(o_{i,t} \mid q,o_{i,<t})}.
\label{reward_computation}
\end{equation}

\subsection{Diagnosing the Perceptual Shortcuts} \label{subsec:diagnostic}

While GRPO (Eq.~\ref{eq:grpoloss}) has achieved superior success in complex reasoning domains like mathematics~\cite{guo2025deepseek}, existing work~\cite{zhang2025tinyllava, wang2025videorft} reveals that its effectiveness on 3B models for video reasoning is surprisingly insignificant. This stark contrast led us to suspect the data itself. Unlike the logical purity of math, we hypothesize that the video reasoning training set possesses a fundamental limitation that causes this failure. This stark contrast led us to suspect the data itself. Unlike the logical purity of math, we hypothesize that the video reasoning training set possesses a fundamental limitation that causes this failure.

To formally test this hypothesis, we designed a diagnostic experiment. We selected the counterfactual task in the CLEVRER~\cite{yi2019clevrer} dataset for this diagnosis, as it is a widely-used benchmark for video reasoning training~\cite{videor1}. Crucially, its detailed collision event annotations (as shown in Figure~\ref{fig:intro}) provide the precise, ground-truth information necessary to formally categorize its counterfactual questions into two distinct types: \textbf{Inferential} and \textbf{Observational}. As illustrated in Figure~\ref{fig:diagnostic_chart}, a truly \textbf{Inferential} question (e.g., Option A) involves a causal blocker. In the video, the ``green, rubber, cube'' prevents the ``yellow, rubber, cylinder'' from colliding with the ``green, metal, sphere.'' To correctly answer ``if the cube is removed,'' the model is forced to perform step-by-step reasoning to deduce a new event that conflicts with the original video's timeline. In contrast, an \textbf{Observational} question (e.g., Option B) can be ``cheated.'' A model does not need to reason about the cube's removal; it can simply scan the original video, observe that the ``purple, rubber, sphere'' and the ``green, metal, sphere'' never interact, and use this superficial visual evidence as a shortcut. This exact mechanism is the ``perceptual shortcut'' we aim to diagnose.

Moreover, this categorization led to a staggering discovery: these ``easy'' pseudo-reasoning Observational shortcuts constitute a massive 74.0\% (13674/18473) of the training dataset. We believe this overwhelming imbalance is the primary culprit. We then fine-tuned 3B and 7B models~\cite {bai2025qwen2} using the GRPO framework on the 1k training samples containing both of these question types, with settings group size $G$=8, $\beta$=0.05, and learning rate $10^{-6}$. 

After training, we evaluate it on the CLEVRER valid dataset, consisting of 11524 observational questions and only 1224 inferential questions. The results were striking. As shown in Figure~\ref{fig:intro}, the 3B model's accuracy on the critical Inferential tasks plummeted from 73.9\% to 63.1\%, even while its performance on Observational tasks remained stable or increased. This provides concrete evidence for our hypothesis: the RL model, poisoned by the shortcut bias, actively unlearned true reasoning.

\begin{figure*}[!t]
    \centering
    \includegraphics[width=1.0\linewidth]{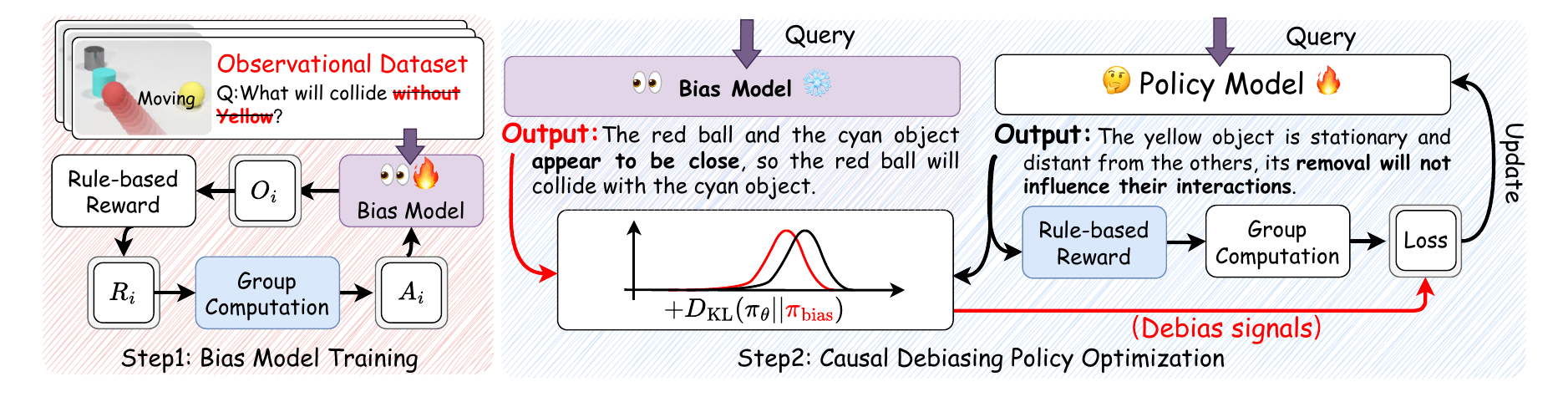}
    \caption{An overview of our VideoThinker framework designed to mitigate perceptual shortcuts in reasoning. We first train a Bias Model to embody those shortcuts. Then CDPO compels the main policy model to learn robust reasoning by using a repulsive KL-divergence objective. This objective actively pushes the policy away from bias.}
    \label{fig:overview}
    \vspace{-5mm}
\end{figure*}

\begin{figure}[!t]
    \centering
    \includegraphics[width=1.0\linewidth]{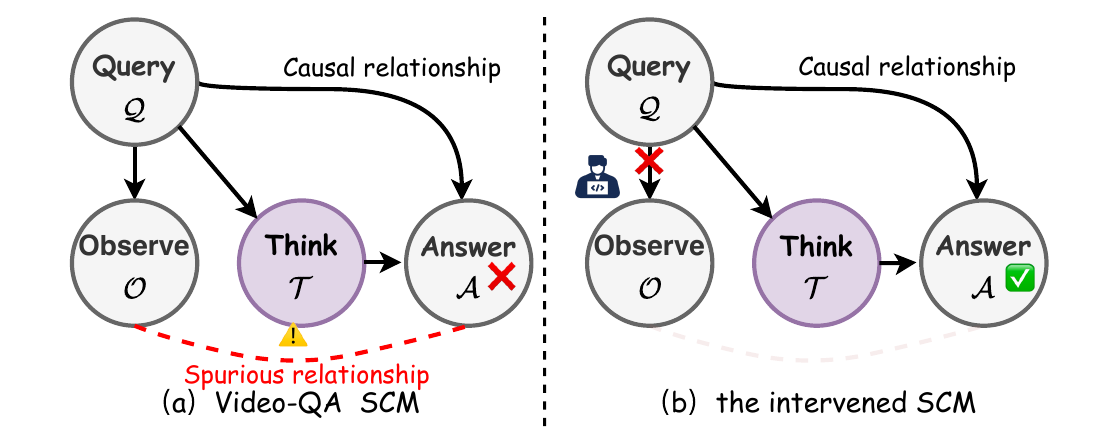}
    \caption{Our modeling. (a) shows our causal assumption in the format of SCM, specifically proposed for the task of Video-QA. (b) shows our method how to cut the perception shortcut, motivate to genuine reasoning.}
    \label{fig:scm}
    \vspace{-5mm}
\end{figure}

\subsection{Causal Analysis of Perceptual Shortcuts}
\label{sec:scm}

We start by formalizing the reasoning process through an SCM~\cite{pearl2009causal}, which explicitly describes the causal relationships between key variables involved in VQA: the input query $\mathcal{Q}$, latent thinking $\mathcal{T}$ reflecting genuine causal reasoning, superficial observation $\mathcal{O}$ capturing statistical biases, and the answer $\mathcal{A}$. This SCM, illustrated in Figure~\ref{fig:scm}, represents our causal assumptions underlying the VQA process:

\begin{itemize}[leftmargin=*, topsep=3pt, itemsep=3pt]
    \item $\mathcal{Q} \rightarrow \mathcal{O}$: The input query $\mathcal{Q}$ is processed to extract superficial, observational evidence, forming the observation variable $\mathcal{O}$. This path represents a shallow form of pattern matching, where the model finds literal events or objects in the video that correspond to the entities in the question, while ignoring complex logical or causal operators.  As illustrated in Figure~\ref{fig:scm}(a), given the query ``What will collide without Yellow?'', this pathway simply extracts the entire observed collision sequence (Red $\rightarrow$ Cyan $\rightarrow$ Metal in Figure~\ref{fig:intro}), as it provides a direct, superficial answer that sidesteps the counterfactual nature of the query.

    \item $\mathcal{Q} \rightarrow \mathcal{T}$: The query $\mathcal{Q}$ is processed to extract information required for genuine reasoning, forming the thinking variable $\mathcal{T}$. For the query ``What will collide without Cyan?'' in the Figure~\ref{fig:scm}(a), this pathway understands that it must simulate an unobserved scenario, leading to the correct inferential chain (Red $\rightarrow$ Metal in Figure~\ref{fig:intro}).

    \item $\mathcal{T} \rightarrow A$: The thinking variable $\mathcal{T}$ causally determines the final answer $\mathcal{A}$. The full chain $\mathcal{Q} \rightarrow \mathcal{T} \rightarrow A$ constitutes the ideal causal pathway, representing genuine reasoning.

    \item $\mathcal{O} \rightarrow \mathcal{A}$: The observation $\mathcal{O}$ can directly influence the answer $\mathcal{A}$. This represents the spurious bias pathway. It is a cognitive shortcut where the model uses the easily extracted observational evidence from $\mathcal{O}$ to directly generate an answer. This path is often exploited by models during training due to its statistical efficiency, but it fails on complex problems requiring true understanding.
\end{itemize}

The SCM reveals a critical issue. The input query $\mathcal{Q}$ acts as a confounder~\cite{pearl2009causal}, creating two parallel information streams. This structure opens a backdoor~\cite{pearl2009causal} between the reasoning process $\mathcal{T}$ and the answer $\mathcal{A}$ via the observation $\mathcal{O}$, leading to spurious correlations. These spurious correlations are a direct result of data bias: the training set is replete with ``easy'' instances where the answer $\mathcal{A}$ can be correctly predicted from superficial observations $\mathcal{O}$ alone, bypassing the need for genuine reasoning $\mathcal{T}$. Consequently, a standard model will inevitably learn to exploit this ``path of least resistance,'' the bias pathway $\mathcal{Q} \rightarrow \mathcal{O} \rightarrow \mathcal{A}$, which compromises its reasoning ability. Therefore, to ensure the model relies on the intended reasoning path $\mathcal{Q} \rightarrow \mathcal{T} \rightarrow \mathcal{A}$, we must intervene to remove the confounding bias rooted in this data, which motivates us to find a solution to debias.

\subsection{The framework of VideoThinker}
\label{causal_debiasing_policy_optimization}

Theoretically, the standard solution from causal inference to eliminate such confounding effects is the backdoor adjustment~\cite{pearl1993bayesian,greenland1999causal}.  This procedure requires calculating the true causal effect by marginalizing over the confounding variable $\mathcal{O}$, conditioned on the context $\mathcal{Q}$:
\begin{equation}
\begin{aligned}
&P(\mathcal{A} | do(\mathcal{T}=t), \mathcal{Q}=q) =
\\ &\int_{\mathcal{O}} P(\mathcal{A} | \mathcal{T}=t, \mathcal{O}=o, \mathcal{Q}=q) P(\mathcal{O}=o | \mathcal{Q}=q) d\mathcal{O},
\end{aligned}
\end{equation}
where this is the bias-free, counterfactual answer. However, this ideal intervention is intractable in our setting. This intractability is twofold: (1) The variable $\mathcal{O}$ is a high-dimensional, continuous latent representation, making the integration over all its possible states computationally prohibitive. (2) The procedure also requires modeling the complex conditional prior $P(\mathcal{O} | \mathcal{Q})$, which is itself a challenging generative task. Therefore, to bridge the gap between theory and practice, we propose \textbf{VideoThinker}, a novel and practical framework for causal debiasing. Instead of attempting the intractable marginalization, VideoThinker achieves the goal of closing the back-door path through an efficient, adversarial approximation. 

\subsubsection{Bias Aware Training}
\label{sec:bias_model_sec}

The first step in our causal intervention framework is to explicitly isolate and embody the ``bad'' bias pathway ($\mathcal{Q} \rightarrow \mathcal{O} \rightarrow \mathcal{A}$). We achieve this by training a dedicated {Bias Model}, denoted as $\pi_{\text{bias}}$, designed to become an expert at exploiting statistical shortcuts. To this end, we leverage the richly annotated CLEVRER~\cite{yi2019clevrer} dataset to construct a specialized bias-promoting dataset, $\mathcal{D}_{\text{bias}}$. We identify observational or shortcut questions within the dataset's counterfactual tasks. These are instances where the counterfactual condition (e.g., ``without the yellow ball'') is irrelevant because the events described in the correct answer choice occur in the original video regardless. Specifically, thanks to the detailed ground-truth collision logs in CLEVRER, we can automatically filter for these samples. For a given question, if an answer option describes an event that perfectly matches an event in the ground-truth log, we classify this instance as observational and add it to the bias dataset, $\mathcal{D}_{\text{bias}}$.

With this curated data, we then aim to accelerate the learning of these shortcut paths. Inspired by DAPO~\citep{yu2025dapo}, we remove the KL-divergence constraint entirely. This encourages the policy to converge quickly to the simplest and most biased solution. 

By optimizing this objective on our curated data, the resulting policy $\pi_{\text{bias}}$ becomes a proficient proxy for the undesirable bias, setting the stage for our causal intervention.

\subsubsection{Causal Debiasing Policy Optimization}
\label{sec:intervention_sec}

With the frozen Bias Model $\pi_{\text{bias}}$ serving as an explicit proxy for the ``bad'' shortcut reasoning, we now train our thinking model, $\pi_{\theta}$. The objective for this model as follows: 
\begin{equation}
\label{eq:cdpoloss}
\begin{aligned}
\mathcal{J}_\text{CDPO}(\theta) =& \mathbb{E}_{(q,a)\sim \mathcal{D}, \{o_i\}_{i=1}^G\sim \pi_{\theta_\text{old}}(\cdot\mid q)} \\&
\Bigg[ \frac{1}{G}\sum_{i=1}^{G} \frac{1}{|o_i|}\sum_{t=1}^{|o_i|} \Bigg( 
\min \Big( r_{i,t}(\theta) \hat{A}_{i,t}, \\&
\ \text{clip} \Big( r_{i,t}(\theta), 1 - \varepsilon, 1 + \varepsilon \Big) \hat{A}_{i,t} \Big)
\\&+ \beta D_{\text{KL}}(\pi_{\theta} \ || \ \pi_{\text{bias}})
\Bigg) \Bigg].
\end{aligned}
\end{equation}
This objective function consists of two key components. It drives $\pi_{\theta}$ to maximize the task-specific advantage $\hat{A}_{i,t}$ on the full dataset $\mathcal{D}$, ensuring the model remains focused on generating correct and high-quality answers. The second term, $+ \beta D_{\text{KL}}(\cdot)$, is our key intervention mechanism. Since the entire $\mathcal{J}_\text{CDPO}(\theta)$ objective is maximized during training, the positive sign before the KL-divergence term means that we are actively training the model to maximize its distributional distance from the frozen bias model. This adversarial pressure penalizes the thinking model for adopting action distributions similar to the shortcut solutions learned by $\pi_{\text{bias}}$. It thereby compels the model to explore and converge upon alternative, more complex reasoning pathways ($\mathcal{Q} \rightarrow \mathcal{T} \rightarrow \mathcal{A}$) that are inaccessible to the bias Model.

In essence, by simultaneously optimizing for task reward and divergence from the bias policy, the VideoThinker framework trains a model that is both accurate and causally robust. The hyperparameter $\beta$ controls the strength of this debiasing regularizer. This entire procedure serves as our practical, gradient-based approximation of the formal backdoor adjustment, effectively cutting the spurious pathway.
\begin{table*}[!t]
\small
\centering
\label{combined_table}
\caption{Comparison of model performance on both video reasoning and general video understanding benchmarks. Following previous works, we restrict evaluation on MMVU$\rm_{mc}$ to multiple-choice questions and exclude subtitles from VideoMME$\rm{_{wo\_{sub}}}$. For CLEVRER$\rm_{cf}$, to mitigate type bias, we specifically select the single-choice counterfactual task for evaluation.}
\vspace{-2mm}
\setlength{\tabcolsep}{0.8mm}
\renewcommand\arraystretch{0.9}
\begin{tabular}{@{}lccccccc@{}}
\toprule
\multirow{2}{*}{\textbf{\textit{Models}}} & \multirow{2}{*}{\textbf{\textit{Training}}} & \multicolumn{3}{c}{\textbf{\textit{Video Reasoning}}} & \multicolumn{3}{c}{{\textbf{\textit{Video Understanding}}}} \\ 
\cmidrule(l){3-5}\cmidrule(l){6-8} 
 &  & CLEVRER$\rm_{cf}$ & MMVU$\rm _{mc}$ & VideoHolmes & MVBench & TempCompass & VideoMME$\rm{_{wo\_{sub}}}$ \\ 
\midrule
\multicolumn{8}{l}{$\bullet$ \textbf{\textit{Open-Source Models}}} \\
LLaMA-VID \cite{li2024llama}      & -  & -          & -  & -       & 41.9& 45.6    & -              \\
VideoLLaMA2 \cite{cheng2024videollama}    & -       & -    & 44.8  & -   & 54.6& -       & 47.9           \\
LongVA-7B \cite{zhang2024long}       & -  & -  & -      & -     & -   & 56.9    & 52.6           \\
VILA-1.5-8B \cite{lin2023vila}     & -  & -  & -        & -   & -   & 58.8    & -              \\
Video-UTR-7B \cite{yu2025unhackable}   & -  & -        & -   & -        & 58.8& 59.7    & 52.6           \\
Qwen2.5-VL-3B (CoT)~\cite{bai2025qwen2} & - & 44.7 & 52.8 & 32.5 & 49.6 & 30.0 & 52.0 \\
\midrule
\multicolumn{8}{l}{$\bullet$ \textbf{\textit{Concurrent R1-based 3B Models}}} \\
TinyLLaVA-Video-R1~\cite{zhang2025tinyllava} & 5.5K SFT\&RL & -  & 46.9 & - & - & 49.5 & 46.6 \\
VideoRFT~\cite{wang2025videorft} & 110K SFT\&RL & 59.3  & 55.1 & 33.0 & 59.5 & 61.0 & 45.4 \\
{Qwen2.5-VL-GRPO}~\cite{shao2024deepseekmath} & 1K RL & {64.9}  & {52.0} & 32.3 & {54.9} & {41.4} & 50.3 \\
{\textbf{VideoThinker-R1}} & 1K RL & {\textbf{79.1}}  & {\textbf{56.8}} &\textbf{ 34.3} & {\textbf{60.9}} & {\textbf{63.5}} & \textbf{52.4} \\
\toprule
\end{tabular}
\label{tab:main_results}
\vspace{-5mm}
\end{table*}

\begin{table}[!t]
\small
\centering
\caption{Ablation study on VideoThinker Framework}
\vspace{-2mm}
\renewcommand{\arraystretch}{0.9}
\setlength{\tabcolsep}{0.5mm}
\begin{tabular}{lcc}
\toprule
\textbf{\textit{Setting}} & {CLEVRER$\rm_{cf}$} & {MMVU$\rm _{mc}$} \\
\midrule
\multicolumn{3}{l}{$\bullet$\textit{\textbf{Reference Model in VideoThinker}}} \\
GRPO (Baseline) & 64.9 & 52.0 \\
VideoThinker-R1 (w/ Qwen2.5VL-3B) & 63.3 & 53.3 \\
VideoThinker-R1 (w/ VideoRFT-3B) & 75.4 & 53.1 \\ 
\rowcolor{gray!5}
VideoThinker-R1 (w/ Bias Model) & \textbf{79.1} & \textbf{56.8} \\
\midrule
\multicolumn{3}{l}{$\bullet$\textit{\textbf{Debiasing Mechanism in VideoThinker}}} \\
KL-Minimization ($- \beta$) & 74.3 & 52.6 \\
\rowcolor{gray!5}
KL-Maximization ($+ \beta$) & \textbf{79.1} & \textbf{56.8} \\
\midrule
\multicolumn{3}{l}{$\bullet$\textit{\textbf{Sensitivity Analysis on $\beta$}}} \\
0.1 & 62.2 & \textbf{58.4} \\
\rowcolor{gray!5}
0.01 & \textbf{79.1} & {56.8} \\
0.001 & 72.8 & 55.5 \\
\bottomrule
\end{tabular}
\label{tab:ablation_components}
\vspace{-6mm}
\end{table}

\section{Experiment}
\subsection{Implementation Details}
We use Qwen2.5-VL-3B-Instruct~\cite{bai2025qwen2} as our base model and conduct training on two NVIDIA RTX A6000 GPUs, each with 48GB VRAM. Training is conducted in two phases: first, the Bias Model is trained for 500 steps on the curated CLEVRER bias dataset. Subsequently, VideoThinker is trained for 500 steps on the CLEVRER training set, with the debiasing coefficient $\beta$ defined in Equation~\ref{eq:cdpoloss} is 0.01. During optimization, we employ a soft accuracy reward and a format reward~\cite{dang2025reinforcing}. To expedite training, we sample up to 16 frames per video and resize frames to a resolution of $128\times28\times28$, with $G$=8 and learning rate $10^{-6}$.

For evaluating, we conduct experiments on six standard benchmarks, consistent with previous work~\cite{videor1,dang2025reinforcing,wang2025videorft}: CLEVRER~\cite{yi2019clevrer}, MMVU~\cite{zhao2025mmvu}, Video-Holmes~\cite{cheng2025video}, MVBench~\cite{li2024mvbench}, TempCompass~\cite{liu2024tempcompass}, and VideoMME~\cite{fu2024video}. Among them, the first three are video reasoning benchmarks, which focus primarily on assessing the model’s reasoning capabilities. The latter three are general-purpose video understanding benchmarks, which include a mixture of perception and reasoning tasks. For all evaluations, we use 32 frames, upscale the input resolution to $256\times28\times28$, and set sampling parameters with top$_p$=0.001 and temperature=0.01. The evaluation metric is average accuracy. More details are provided in the supplementary.

\subsection{Comparison with State-of-the-Arts}
As shown in Table~\ref{tab:main_results}, our proposed causal intervention framework, VideoThinker, empowers VideoThinker-R1 to achieve comprehensive leadership as a lightweight 3B model, even outperforming several larger 7B-scale open-source models. For instance, on general understanding benchmarks like MVBench and TempCompass, our model's scores of 60.9 and 63.5 surpass those of Video-UTR-7B~\cite{yu2025unhackable}, showcasing the exceptional performance achievable with our efficient approach. Specifically, on the CLEVRER benchmark, VideoThinker-R1 achieves a score of 79.1\%, outperforming GRPO by a significant margin of 14.2\% under identical training conditions. This highlights that the performance leap is driven not by data differences but by VideoThinker's ability to counteract shortcut learning. Furthermore, the improved generalization fostered by VideoThinker is evident when compared to other contemporary 3B models like VideoRFT-3B~\cite{wang2025videorft} and TinyLLaVA-Video-R1~\cite{zhang2025tinyllava}. Our model demonstrates holistic improvements, leading on MVBench and TempCompass, with an even more pronounced advantage on VideoMME where it surpasses VideoRFT-3B by 7.0\%. These results across diverse tasks validate that VideoThinker provides an effective framework for enhancing the general reasoning abilities.

\subsection{Ablation Study}

\subsubsection{Choice of Bias Model}
A core premise of VideoThinker is that effective debiasing requires a reference model that accurately represents the spurious pathway. We first establish baseline performances under the GRPO framework, which uses a model's own initial policy as a reference to constrain fine-tuning. Subsequently, we evaluate the effect of using different external models as the ``repulsive target'' for VideoThinker. As shown in Table~\ref{tab:ablation_components}, within the VideoThinker framework, employing the purpose-built {Bias Model}, a model specifically designed to embody shortcut behaviors, as the repulsive target yields the best performance, surging to {79.1\%} on CLEVRER$_{\text{cf}}$ and {56.8\%} on MMVU$_{\text{mc}}$. This result significantly outperforms variants that repel a best model (like VideoRFT-3B) or self-repulsion (which leads to policy collapse). This provides strong evidence for our premise.
\begin{figure}[!t]
    \centering
    \includegraphics[width=1.0\linewidth]{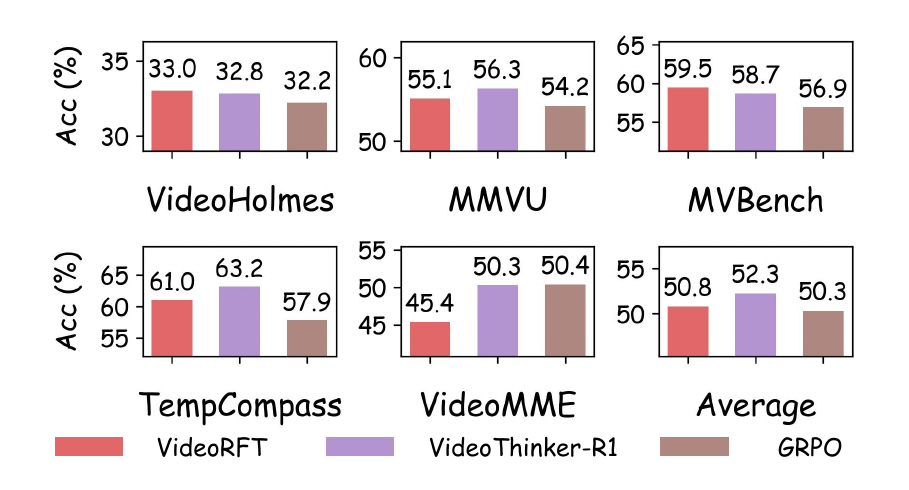}
    \caption{Performance of our method training on other datasets.}
    \label{fig:perceptiontest}
    \vspace{-6mm}
\end{figure}
\begin{figure*}[!t]
    \centering
    \includegraphics[width=1.0\linewidth]{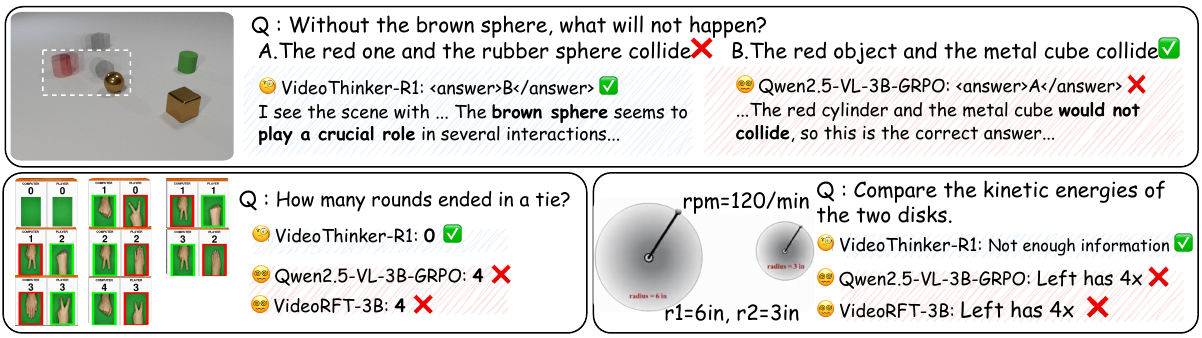}
    \vspace{-4mm}
    \caption{Qualitative examples of VideoThinker-R1's reasoning performance. 
(\textbf{Top}) For a complex reasoning from the CLEVRER dataset, we visualize the detailed reasoning path generated by the model to reach the correct answer. (\textbf{Bottom}) For two different queries from the MMVU dataset, we display the model's correct final answers.}
    \label{fig:sample}
    \vspace{-6mm}
\end{figure*}

\subsubsection{The Repulsive Debiasing Mechanism}
Next, we validate the directionality of our intervention: pushing away from (KL-Maximization) or pulling towards (KL-Minimization) the Bias Model's policy. Table~\ref{tab:ablation_components} shows KL-Minimization (attraction) fails to generalize, yielding no improvement on MMVU. In contrast, our proposed KL-Maximization (repulsion) achieves superior results, confirming that robust reasoning emerges not from imitating flawed patterns, but from being explicitly repelled by them. Having confirmed repulsion, we analyze the sensitivity of its strength, $\beta$ (Table~\ref{tab:ablation_components}). The results reveal a nuanced trade-off. Performance on the in-domain CLEVRER benchmark is sensitive to $\beta$. This is expected, as an overly large $\beta$ (e.g., 0.1) can over-penalize, forcing the model to unlearn useful foundational knowledge (like collision logic) along with the shortcuts. In sharp contrast, performance on the general-purpose MMVU benchmark is highly robust, showing little variation. This indicates that the generalizable reasoning capability learned by our model is stable and insensitive to $\beta$ within this range. As $\beta=0.01$ achieves the peak CLEVRER score while preserving this strong generalization, we adopt it for all main experiments.

\subsubsection{Robustness to Train Data}
Moreover, we assess whether VideoThinker's effectiveness is limited to datasets rich in counterfactual queries. To this end, we conduct experiments on other training dataset~\cite{patraucean2023perception}, which only has 1.9\% inferential problems. As shown in Figure~\ref{fig:perceptiontest}, our VideoThinker-R1 achieves an average accuracy of 52.3\%, surpassing the GRPO baseline (50.3\%) and the stronger VideoRFT-3B (50.8\%). This result is particularly compelling because it shows that VideoThinker is not merely overfitting to a specific problem type. Instead, our framework enhances the model's fundamental reasoning by teaching it to isolate and suppress spurious signals, a skill that is effective even when counterfactual queries are sparse.

\subsubsection{Scaling to 7B params}
Finally, we extend our evaluation to the 7B scale to assess the framework's scalability (Table~\ref{tab:7b}). While achieving results comparable to SOTA 7B models~\cite{videor1,wang2025videorft}, the gains are more direct on the 3B model. This is because the ``perceptual shortcut'' issue is primarily a deficit of limited-capacity models. Since larger models are inherently more robust against learning bias (Figure~\ref{fig:intro}), our method is particularly critical for enabling compact models to overcome these bottlenecks and achieve robust reasoning.

\begin{table}[!t]
\small
\centering
\caption{Performance comparison at the 7B parameter scale}
\vspace{-2mm}
\renewcommand{\arraystretch}{0.9}
\setlength{\tabcolsep}{4mm}
\begin{tabular}{lcc}
\toprule
\textbf{\textit{Models}} & MVBench & TempCompass \\
\midrule
Video-R1-7B~\cite{videor1} & 63.8 & {73.2} \\
VideoRFT-7B~\cite{wang2025videorft} & {62.1} & \textbf{73.7} \\
\rowcolor{gray!5}
VideoThinker-R1-7B & \textbf{65.0} & \textbf{73.7} \\
\bottomrule
\end{tabular}
\label{tab:7b}
\vspace{-5mm}
\end{table}

\subsection{Qualitative Analysis}
To qualitatively illustrate how VideoThinker mitigates shortcut learning, we analyze several reasoning tasks from the CLEVRER and MMVU datasets. As shown in Figure~\ref{fig:sample}, the baseline model exhibits a reliance on cognitive shortcuts. For instance, in the MMVU game scenarios, the model often defaults to an incorrect decision based on the superficial visual cue of ``identical scores,'' bypassing a necessary comprehension of the game's rules. This tendency is more pronounced in the CLEVRER dataset. Here, the baseline model generates a plausible-seeming textual rationale, yet provides a final answer that directly conflicts with this statement. This validates our core hypothesis that the baseline model does not perform genuine logical reasoning. Instead, it over-relies on surface-level cues, resulting in a fractured logical chain.

Conversely, VideoThinker-R1 effectively overcomes this limitation. In the CLEVRER task, the model correctly identifies the ``brown sphere'' as the critical object through logical analysis, leading to a correct final decision and demonstrating a complete and coherent reasoning process. Similarly, in the MMVU task, VideoThinker-R1 is not misled by superficial information like ``identical scores'' and instead renders a judgment grounded in the game's rules. When faced with an ambiguous turntable momentum problem with incomplete information, the model correctly concludes that the question is unanswerable given the existing data. These examples demonstrate that our proposed method effectively suppresses the model's reliance on spurious features and enforces the desired logical reasoning.

\section{Conclusions}
In this work, we diagnosed perceptual bias as a key factor limiting the reasoning abilities of lightweight MLLMs and proposed VideoThinker to counteract it. Our framework trains a ``bias model'' to master ``perceptual shortcuts'' and then employs an innovative repulsive objective that forces our primary model to discover genuine reasoning pathways. Experimental results across six benchmarks validate the effectiveness of our approach. We hope this work provides a foundation for building more generalized MLLMs.
\newpage
\section*{Acknowledgments}
This work was supported by the National Natural Science Foundation of China under Grant No.62506393, the Guangdong Basic and Applied Basic Research Foundation under Grant No.2026A1515011438, and the Post- doctoral Fellowship Program and the China Postdoctoral Science Foundation under Grant No.GZC20252314.
{
    \small
    \bibliographystyle{unsrt}
    \bibliography{main}

\begin{thebibliography}{10}

\bibitem{shao2024deepseekmath}
Zhihong Shao, Peiyi Wang, Qihao Zhu, Runxin Xu, Junxiao Song, Xiao Bi, Haowei Zhang, Mingchuan Zhang, YK~Li, Y~Wu, et~al.
\newblock Deepseekmath: Pushing the limits of mathematical reasoning in open language models.
\newblock {\em arXiv preprint arXiv:2402.03300}, 2024.

\bibitem{xiao2025fastslow}
Wenyi Xiao and Leilei Gan.
\newblock Fast-slow thinking {GRPO} for large vision-language model reasoning.
\newblock In {\em NeurIPS}, 2025.

\bibitem{wang2025timer}
Ye~Wang, Ziheng Wang, Boshen Xu, Yang Du, Kejun Lin, Zihan Xiao, Zihao Yue, Jianzhong Ju, Liang Zhang, Dingyi Yang, Xiangnan Fang, Zewen He, Zhenbo Luo, Wenxuan Wang, Junqi Lin, Jian Luan, and Qin Jin.
\newblock Time-r1: Post-training large vision language model for temporal video grounding.
\newblock In {\em NeurIPS}, 2025.

\bibitem{videor1}
Kaituo Feng, Kaixiong Gong, Bohao Li, Zonghao Guo, Yibing Wang, Tianshuo Peng, Junfei Wu, Xiaoying Zhang, Benyou Wang, and Xiangyu Yue.
\newblock Video-r1: Reinforcing video reasoning in {MLLM}s.
\newblock In {\em NeurIPS}, 2025.

\bibitem{li2025videochat}
Xinhao Li, Ziang Yan, Desen Meng, Lu~Dong, Xiangyu Zeng, Yinan He, Yali Wang, Yu~Qiao, Yi~Wang, and Limin Wang.
\newblock Videochat-r1: Enhancing spatio-temporal perception via reinforcement fine-tuning.
\newblock {\em arXiv preprint arXiv:2504.06958}, 2025.

\bibitem{dang2025reinforcing}
Jisheng Dang, Jingze Wu, Teng Wang, Xuanhui Lin, Nannan Zhu, Hongbo Chen, Wei-Shi Zheng, Meng Wang, and Tat-Seng Chua.
\newblock Reinforcing video reasoning with focused thinking.
\newblock {\em arXiv preprint arXiv:2505.24718}, 2025.

\bibitem{zhang2025tinyllava}
Xingjian Zhang, Siwei Wen, Wenjun Wu, and Lei Huang.
\newblock Tinyllava-video-r1: Towards smaller lmms for video reasoning.
\newblock {\em arXiv preprint arXiv:2504.09641}, 2025.

\bibitem{wang2025videorft}
Qi~Wang, Yanrui Yu, Ye~Yuan, Rui Mao, and Tianfei Zhou.
\newblock Video{RFT}: Incentivizing video reasoning capability in {MLLM}s via reinforced fine-tuning.
\newblock In {\em NeurIPS}, 2025.

\bibitem{yi2019clevrer}
Kexin Yi, Chuang Gan, Yunzhu Li, Pushmeet Kohli, Jiajun Wu, Antonio Torralba, and Joshua~B Tenenbaum.
\newblock Clevrer: Collision events for video representation and reasoning.
\newblock In {\em ICLR}, 2020.

\bibitem{chen2024multi}
Xuweiyi Chen, Ziqiao Ma, Xuejun Zhang, Sihan Xu, Shengyi Qian, Jianing Yang, David Fouhey, and Joyce Chai.
\newblock Multi-object hallucination in vision language models.
\newblock In {\em NeurIPS}, volume~37, pages 44393--44418, 2024.

\bibitem{shukor2024beyond}
Mustafa Shukor, Alexandre Rame, Corentin Dancette, and Matthieu Cord.
\newblock Beyond task performance: evaluating and reducing the flaws of large multimodal models with in-context-learning.
\newblock In {\em ICLR}, 2024.

\bibitem{xu2025more}
Zhongxing Xu, Chengzhi Liu, Qingyue Wei, Juncheng Wu, James Zou, Xin~Eric Wang, Yuyin Zhou, and Sheng Liu.
\newblock More thinking, less seeing? assessing amplified hallucination in multimodal reasoning models.
\newblock In {\em NeurIPS}, 2025.

\bibitem{bai2025qwen2}
Shuai Bai, Keqin Chen, Xuejing Liu, Jialin Wang, Wenbin Ge, Sibo Song, Kai Dang, Peng Wang, Shijie Wang, Jun Tang, et~al.
\newblock Qwen2. 5-vl technical report.
\newblock {\em arXiv preprint arXiv:2502.13923}, 2025.

\bibitem{fu2024video}
Chaoyou Fu, Yuhan Dai, Yongdong Luo, Lei Li, Shuhuai Ren, Renrui Zhang, Zihan Wang, Chenyu Zhou, Yunhang Shen, Mengdan Zhang, et~al.
\newblock Video-mme: The first-ever comprehensive evaluation benchmark of multi-modal llms in video analysis.
\newblock In {\em CVPR}, pages 24108--24118, 2025.

\bibitem{zhan2025visionr1}
Yufei Zhan, Yousong Zhu, Shurong Zheng, Hongyin Zhao, Fan Yang, Ming Tang, and Jinqiao Wang.
\newblock Vision-r1: Evolving human-free alignment in large vision-language models via vision-guided reinforcement learning.
\newblock {\em arXiv preprint arXiv:2503.18013}, 2025.

\bibitem{torralba2011unbiased}
Antonio Torralba and Alexei~A Efros.
\newblock Unbiased look at dataset bias.
\newblock In {\em CVPR}, pages 1521--1528, 2011.

\bibitem{zhang2021seeing}
Quan Zhang, Jianhuang Lai, Zhanxiang Feng, and Xiaohua Xie.
\newblock Seeing like a human: Asynchronous learning with dynamic progressive refinement for person re-identification.
\newblock {\em IEEE TIP}, 31:352--365, 2021.

\bibitem{zhang2024separable}
Quan Zhang, Jianhuang Lai, Xiaohua Xie, Xiaofeng Jin, and Sien Huang.
\newblock Separable spatial-temporal residual graph for cloth-changing group re-identification.
\newblock {\em IEEE TPAMI}, 46(8):5791--5805, 2024.

\bibitem{zhang2025debiasing}
Zefeng Zhang, Hengzhu Tang, Jiawei Sheng, Zhenyu Zhang, Yiming Ren, Zhenyang Li, Dawei Yin, Duohe Ma, and Tingwen Liu.
\newblock Debiasing multimodal large language models via noise-aware preference optimization.
\newblock In {\em CVPR}, pages 9423--9433, 2025.

\bibitem{si2022language}
Qingyi Si, Fandong Meng, Mingyu Zheng, Zheng Lin, Yuanxin Liu, Peng Fu, Yanan Cao, Weiping Wang, and Jie Zhou.
\newblock Language prior is not the only shortcut: A benchmark for shortcut learning in vqa.
\newblock In {\em EMNLP Findings}, pages 3698--3712, 2022.

\bibitem{Zang_2023_CVPR}
Chuanqi Zang, Hanqing Wang, Mingtao Pei, and Wei Liang.
\newblock Discovering the real association: Multimodal causal reasoning in video question answering.
\newblock In {\em CVPR}, pages 19027--19036, June 2023.

\bibitem{zhou2024mitigating}
Guanyu Zhou, Yibo Yan, Xin Zou, Kun Wang, Aiwei Liu, and Xuming Hu.
\newblock Mitigating modality prior-induced hallucinations in multimodal large language models via deciphering attention causality.
\newblock In {\em ICLR}, 2025.

\bibitem{park2025assessing}
Jean Park, Kuk~Jin Jang, Basam Alasaly, Sriharsha Mopidevi, Andrew Zolensky, Eric Eaton, Insup Lee, and Kevin Johnson.
\newblock Assessing modality bias in video question answering benchmarks with multimodal large language models.
\newblock In {\em AAAI}, volume~39, pages 19821--19829, 2025.

\bibitem{niu2021counterfactual}
Yulei Niu, Kaihua Tang, Hanwang Zhang, Zhiwu Lu, Xian-Sheng Hua, and Ji-Rong Wen.
\newblock Counterfactual vqa: A cause-effect look at language bias.
\newblock In {\em CVPR}, pages 12700--12710, 2021.

\bibitem{chen-etal-2024-quantifying}
Meiqi Chen, Yixin Cao, Yan Zhang, and Chaochao Lu.
\newblock Quantifying and mitigating unimodal biases in multimodal large language models: A causal perspective.
\newblock In {\em EMNLP Findings}, pages 16449--16469, 2024.

\bibitem{su2023language}
Hung-Ting Su, Yulei Niu, Xudong Lin, Winston~H Hsu, and Shih-Fu Chang.
\newblock Language models are causal knowledge extractors for zero-shot video question answering.
\newblock In {\em CVPR}, pages 4950--4959, 2023.

\bibitem{tai2023link}
Yan Tai, Weichen Fan, Zhao Zhang, and Ziwei Liu.
\newblock Link-context learning for multimodal llms.
\newblock In {\em CVPR}, pages 27176--27185, 2024.

\bibitem{zhao2023causal}
Shitian Zhao, Zhuowan Li, Yadong Lu, Alan Yuille, and Yan Wang.
\newblock Causal-cog: A causal-effect look at context generation for boosting multi-modal language models.
\newblock In {\em CVPR}, pages 13342--13351, 2024.

\bibitem{chen2025cross}
Weixing Chen, Yang Liu, Binglin Chen, Jiandong Su, Yongsen Zheng, and Liang Lin.
\newblock Cross-modal causal relation alignment for video question grounding.
\newblock In {\em CVPR}, pages 24087--24096, 2025.

\bibitem{qin2021causal}
Wei Qin, Hanwang Zhang, Richang Hong, Ee-Peng Lim, and Qianru Sun.
\newblock Causal interventional training for image recognition.
\newblock {\em IEEE TMM}, 25:1033--1044, 2021.

\bibitem{zhang2021deep}
Xingxuan Zhang, Peng Cui, Renzhe Xu, Linjun Zhou, Yue He, and Zheyan Shen.
\newblock Deep stable learning for out-of-distribution generalization.
\newblock In {\em CVPR}, pages 5372--5382, 2021.

\bibitem{zhang2021learning}
Quan Zhang, Jianhuang Lai, and Xiaohua Xie.
\newblock Learning modal-invariant angular metric by cyclic projection network for vis-nir person re-identification.
\newblock {\em IEEE TIP}, 30:8019--8033, 2021.

\bibitem{zhang2023wavelet}
Quan Zhang, Jianhuang Lai, Junyong Zhu, and Xiaohua Xie.
\newblock Wavelet-guided promotion-suppression transformer for surface-defect detection.
\newblock {\em IEEE TIP}, 32:4517--4528, 2023.

\bibitem{zhang2024view}
Quan Zhang, Lei Wang, Vishal~M Patel, Xiaohua Xie, and Jianhaung Lai.
\newblock View-decoupled transformer for person re-identification under aerial-ground camera network.
\newblock In {\em CVPR}, pages 22000--22009, 2024.

\bibitem{hu2025video}
Kairui Hu, Penghao Wu, Fanyi Pu, Wang Xiao, Yuanhan Zhang, Xiang Yue, Bo~Li, and Ziwei Liu.
\newblock Video-mmmu: Evaluating knowledge acquisition from multi-discipline professional videos.
\newblock {\em arXiv preprint arXiv:2501.13826}, 2025.

\bibitem{li2023videochat}
KunChang Li, Yinan He, Yi~Wang, Yizhuo Li, Wenhai Wang, Ping Luo, Yali Wang, Limin Wang, and Yu~Qiao.
\newblock Videochat: Chat-centric video understanding.
\newblock {\em arXiv preprint arXiv:2305.06355}, 2023.

\bibitem{guo2025deepseek}
Daya Guo, Dejian Yang, Haowei Zhang, Junxiao Song, Peiyi Wang, Qihao Zhu, Runxin Xu, Ruoyu Zhang, Shirong Ma, Xiao Bi, et~al.
\newblock Deepseek-r1 incentivizes reasoning in llms through reinforcement learning.
\newblock {\em Nature}, 645(8081):633--638, 2025.

\bibitem{yu2025dapo}
Qiying Yu, Zheng Zhang, Ruofei Zhu, Yufeng Yuan, Xiaochen Zuo, YuYue, Weinan Dai, Tiantian Fan, Gaohong Liu, Juncai Liu, LingJun Liu, Xin Liu, Haibin Lin, Zhiqi Lin, Bole Ma, Guangming Sheng, Yuxuan Tong, Chi Zhang, Mofan Zhang, Ru~Zhang, Wang Zhang, Hang Zhu, Jinhua Zhu, Jiaze Chen, Jiangjie Chen, Chengyi Wang, Hongli Yu, Yuxuan Song, Xiangpeng Wei, Hao Zhou, Jingjing Liu, Wei-Ying Ma, Ya-Qin Zhang, Lin Yan, Yonghui Wu, and Mingxuan Wang.
\newblock {DAPO}: An open-source {LLM} reinforcement learning system at scale.
\newblock In {\em NeurIPS}, 2025.

\bibitem{pearl2009causal}
Judea Pearl.
\newblock Causal inference in statistics: An overview.
\newblock 2009.

\bibitem{pearl1993bayesian}
Judea Pearl.
\newblock [bayesian analysis in expert systems]: Comment: Graphical models, causality and intervention.
\newblock {\em Statistical Science}, 8(3):266--269, 1993.

\bibitem{greenland1999causal}
Sander Greenland, Judea Pearl, and James~M Robins.
\newblock Causal diagrams for epidemiologic research.
\newblock {\em Epidemiology}, 10(1):37--48, 1999.

\bibitem{li2024llama}
Yanwei Li, Chengyao Wang, and Jiaya Jia.
\newblock Llama-vid: An image is worth 2 tokens in large language models.
\newblock In {\em ECCV}, pages 323--340. Springer, 2024.

\bibitem{cheng2024videollama}
Zesen Cheng, Sicong Leng, Hang Zhang, Yifei Xin, Xin Li, Guanzheng Chen, Yongxin Zhu, Wenqi Zhang, Ziyang Luo, Deli Zhao, et~al.
\newblock Videollama 2: Advancing spatial-temporal modeling and audio understanding in video-llms.
\newblock {\em arXiv preprint arXiv:2406.07476}, 2024.

\bibitem{zhang2024long}
Peiyuan Zhang, Kaichen Zhang, Bo~Li, Guangtao Zeng, Jingkang Yang, Yuanhan Zhang, Ziyue Wang, Haoran Tan, Chunyuan Li, and Ziwei Liu.
\newblock Long context transfer from language to vision.
\newblock {\em arXiv preprint arXiv:2406.16852}, 2024.

\bibitem{lin2023vila}
Ji~Lin, Hongxu Yin, Wei Ping, Pavlo Molchanov, Mohammad Shoeybi, and Song Han.
\newblock Vila: On pre-training for visual language models.
\newblock In {\em CVPR}, pages 26689--26699, 2024.

\bibitem{yu2025unhackable}
En~Yu, Kangheng Lin, Liang Zhao, Yana Wei, Zining Zhu, Haoran Wei, Jianjian Sun, Zheng Ge, Xiangyu Zhang, Jingyu Wang, et~al.
\newblock Unhackable temporal rewarding for scalable video mllms.
\newblock {\em arXiv preprint arXiv:2502.12081}, 2025.

\bibitem{zhao2025mmvu}
Yilun Zhao, Haowei Zhang, Lujing Xie, Tongyan Hu, Guo Gan, Yitao Long, Zhiyuan Hu, Weiyuan Chen, Chuhan Li, Zhijian Xu, et~al.
\newblock Mmvu: Measuring expert-level multi-discipline video understanding.
\newblock In {\em CVPR}, pages 8475--8489, 2025.

\bibitem{cheng2025video}
Junhao Cheng, Yuying Ge, Teng Wang, Yixiao Ge, Jing Liao, and Ying Shan.
\newblock Video-holmes: Can mllm think like holmes for complex video reasoning?
\newblock {\em arXiv preprint arXiv:2505.21374}, 2025.

\bibitem{li2024mvbench}
Kunchang Li, Yali Wang, Yinan He, Yizhuo Li, Yi~Wang, Yi~Liu, Zun Wang, Jilan Xu, Guo Chen, Ping Luo, et~al.
\newblock Mvbench: A comprehensive multi-modal video understanding benchmark.
\newblock In {\em CVPR}, pages 22195--22206, 2024.

\bibitem{liu2024tempcompass}
Yuanxin Liu, Shicheng Li, Yi~Liu, Yuxiang Wang, Shuhuai Ren, Lei Li, Sishuo Chen, Xu~Sun, and Lu~Hou.
\newblock Tempcompass: Do video llms really understand videos?
\newblock In {\em ACL Findings}, pages 8731--8772, 2024.

\bibitem{patraucean2023perception}
Viorica Patraucean, Lucas Smaira, Ankush Gupta, Adria~Recasens Continente, Larisa Markeeva, Dylan~Sunil Banarse, Skanda Koppula, Joseph Heyward, Mateusz Malinowski, Yi~Yang, Carl Doersch, Tatiana Matejovicova, Yury Sulsky, Antoine Miech, Alexandre Fr{\'e}chette, Hanna Klimczak, Raphael Koster, Junlin Zhang, Stephanie Winkler, Yusuf Aytar, Simon Osindero, Dima Damen, Andrew Zisserman, and Joao Carreira.
\newblock Perception test: A diagnostic benchmark for multimodal video models.
\newblock In {\em NeurIPS}, 2023.

\end{thebibliography}
}

\newpage
\clearpage
\appendix
\setcounter{page}{1}

\maketitlesupplementary

In this document, we provide additional details on the implementation and benchmarks to complement the main paper. Further experiments are also incorporated. And we provided all the code in the supplement, including the JSON files for bias and VideoThinker-R1 training. Specifically, we introduce the details of the training setting in Sec.~\ref {experiment_details}. And in Sec.~\ref{diagnostic_details}, we illustrate how to filter the inferential and observational data, including the algorithm  and samples, and the details of the setups. Finally, the discussion of the limitations is provided in Sec.~\ref{limitation}. 

\section{Detailed Experimental Setup}
\label{experiment_details}
\subsection{Training Setup}
\paragraph{Prompt.} For training, we use the simple prompting strategy like TW-GRPO~\cite{dang2025reinforcing}, the prompt is:
\textit{``Output the thinking process in $<$think$>$$<$/think$>$ and the final answer (letters separated by commas, if multiple) in $<$answer$><$/answer$>$ tags."}
\paragraph{Reward.}
VideoThinker-R1 uses two types of rewards:
\begin{itemize}[leftmargin=*]
\item \textbf{Format Reward.} Similar to other existing MLLM-R1~\cite{videor1,li2025videochat}, we introduce format rewards to ensure the model outputs responses in the desired format. For example, we expect the model to enclose its thought process within \texttt{<think>...</think>} and the answer within \texttt{<answer>...</answer>}. We design a format reward $R_{\mathrm{format}}$ for each task and use regular expression matching to determine whether the model adheres to the specified format.
\item \textbf{Multi-Level Soft Reward.} To address the high reward variance in complex reasoning tasks with multiple correct answers, we choose a soft reward~\cite{dang2025reinforcing} to provide a more granular learning signal. This reward is designed to assign partial credit for incomplete yet correct predictions while strictly penalizing any false positives. Specifically, the reward $R_{soft}$ is calculated as the ratio of correctly predicted items to the total ground truth items ($|P|/|G|$) if and only if the predicted set $P$ is a subset of the ground truth set $G$. If any prediction is outside the ground truth set, the reward is zero, thus promoting precision. This fine-grained feedback on accuracy leads to more stable gradient estimation and policy optimization.
\end{itemize}

\paragraph{Bias Model Training.} The first step in our VideoThinker framework is to create a specialized Bias Model ($\pi_{bias}$) that explicitly learns and embodies the ``perceptual shortcut" behavior. To construct its training data, we begin with the counterfactual subset of the CLEVRER~\cite{yi2019clevrer} dataset. Using the filtering method detailed in Section~\ref{bias_dataset}, we curate a set of 12,191 (out of 18,473) ``perceptual" samples, where the correct answer can be directly observed from the video. Crucially, to compel the model to adopt a shortcut, we programmatically simplify these questions into purely observational tasks by removing the counterfactual condition. For instance, the question \textit{``Which event will happen if the cylinder is removed?"} is transformed into the simpler \textit{``Which event will happen?"}. This modification forces the model to ignore the reasoning premise and instead learn a policy that describes only visual events, thereby intentionally instilling the desired perceptual bias.

To train the bias model, we randomly selected 500 samples from this curated dataset and fine-tuned the base model for 500 steps. We employed the GRPO~\cite{shao2024deepseekmath} but deliberately removed its KL-divergence constraint to accelerate the model's convergence to the biased policy~\cite{yu2025dapo}. Other training parameters, such as a learning rate of $10^{-6}$, follow established work. On a single NVIDIA RTX A6000 GPU with 48GB VRAM, this fine-tuning process takes approximately 4.5 hours to complete.

\paragraph{VideoThinker-R1 Training.} With the frozen Bias Model ($\pi_{bias}$), we proceed to fine-tune our primary reasoning model, VideoThinker-R1. For this stage, we randomly sample 1,000 examples from the original CLEVRER~\cite{yi2019clevrer} counterfactual training set. The model is trained for 500 steps using CDPO, as defined in Equation~\ref{eq:cdpoloss}. In this objective, the crucial hyperparameter $\beta$, which controls the strength of the repulsive force against the bias model, is set to 0.001. On two NVIDIA RTX A6000 GPUs with 48GB VRAM, this fine-tuning process takes approximately 6 hours to complete.

\subsection{Evaluating Setup}

\label{evaluation_setting}
Our model's performance is assessed across six diverse video benchmarks, which we categorize into two groups to ensure a comprehensive evaluation. The first group, focused on general video understanding, includes MVBench \cite{li2024mvbench}, TempCompass \cite{liu2024tempcompass}, and VideoMME \cite{fu2024video}. These benchmarks primarily test core visual perception and temporal comprehension abilities. The second group is designed to evaluate complex reasoning, featuring CLEVRER \cite{yi2019clevrer}, Video-Holmes \cite{cheng2025video}, and MMVU \cite{zhao2025mmvu}. These datasets assess sophisticated spatiotemporal and multimodal reasoning over dynamic video content.

For all evaluations, we follow the experimental setup used in Video-RFT \cite{wang2025videorft}, using identical prompts, sampling temperature (0.01), top\_p (0.001), and batch size to ensure consistency. For CLEVRER~\cite{yi2019clevrer}, following the work~\cite{dang2025reinforcing}, we evaluate exclusively on its most challenging counterfactual subset. To maintain a fair comparison with models that do not support a multiple-answer format, we adopt a single-answer evaluation subset. For other benchmarks, we align with the setup in Video-RFT \cite{wang2025videorft}, conducting experiments on excluding subtitles from VideoMME~\cite{fu2024video} and the multiple-choice split of MMVU~\cite{zhao2025mmvu}.

\section{Details of Diagnostic Experiment}
\label{diagnostic_details}
\renewcommand{\thefigure}{A\arabic{figure}}
\setcounter{figure}{0}
\renewcommand{\thetable}{A\arabic{table}}
\setcounter{table}{0}
\begin{figure*}[!t]
    \centering
    \includegraphics[width=1.0\linewidth]{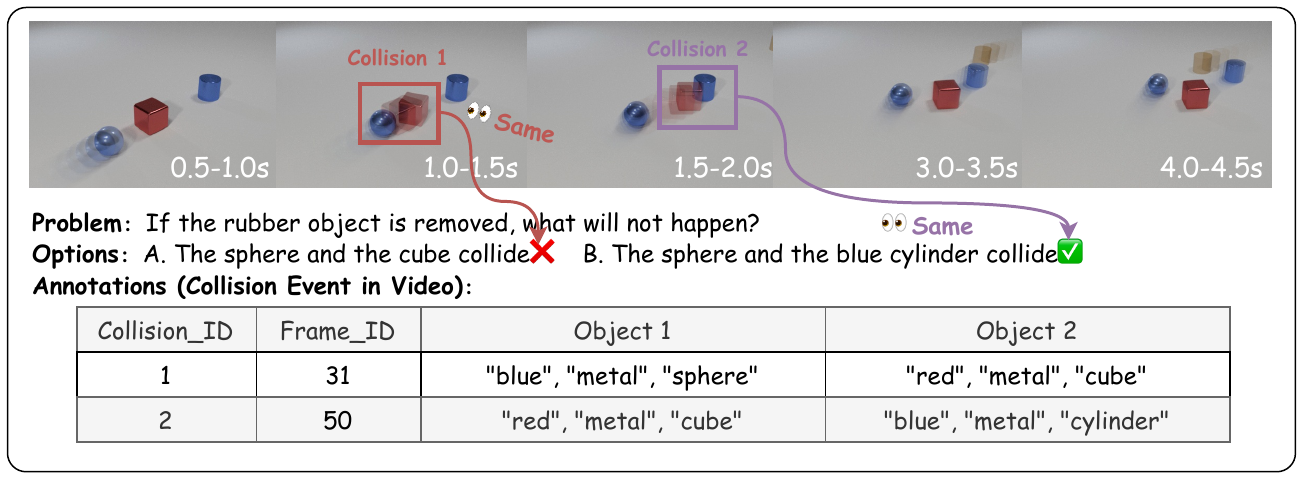}
    \caption{The example of the Observational Problem in CLEVRER.}
    \label{fig:obs}
\end{figure*}

\begin{figure*}[!t]
    \centering
    \includegraphics[width=1.0\linewidth]{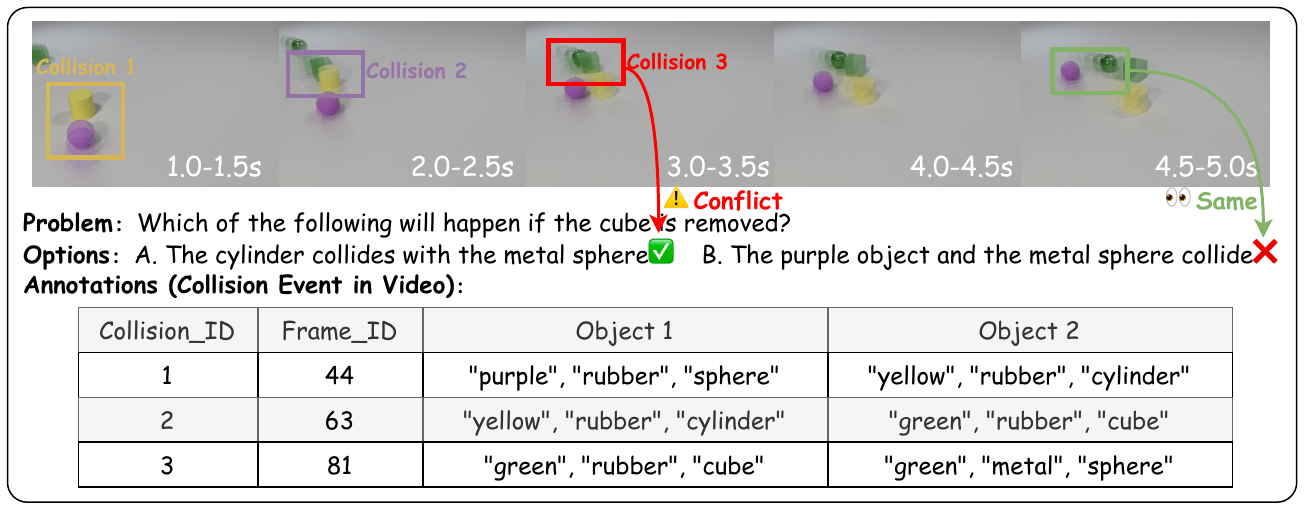}
    \caption{The example of the Inferential Problem in CLEVRER.}
    \label{fig:infer}
\end{figure*}

To better understand the root causes of performance degradation in smaller models fine-tuned under perceptual biases, we constructed a diagnostic experiment that disentangles two qualitatively different reasoning types: \textbf{observational} and \textbf{inferential}. This diagnostic task was motivated by the observation that perceptual shortcuts, heuristics that exploit surface-level correlations in visual outputs, often suffice for solving a subset of questions, while failing for others that require causal or counterfactual inference. To concretely illustrate this distinction, we curated visual question-answering datasets from the CLEVRER benchmark~\cite{yi2019clevrer} and manually annotated questions into two categories:

\begin{itemize}[leftmargin=*]
\item \textbf{Observational questions}, where a model succeeds by detecting and describing what visibly occurred in the video.

\item \textbf{Inferential questions}, where solving the task requires reasoning about hypothetical interventions.
\end{itemize}

In the following sections, we present detailed examples from each category to clarify their defining characteristics and implications for model behavior. We also provide a formal description of our problem filtering strategy, which employs a rule-based classification algorithm to automatically distinguish between observational and inferential questions, thereby enabling large-scale analysis of capability conflicts under different fine-tuning regimes.

\subsection{Examples of the Observational Problem}
Observational questions are characterized by the fact that the correct answer can be derived directly from what is visually present in the video. Figure~\ref{fig:obs} presents a representative example of an \textit{observational} question from the CLEVRER dataset. In this case, the question asks what event will not happen if the rubber object is removed. Among the candidate options, ``the sphere and the blue cylinder collide" corresponds to a visible event in the original video, and this event remains unaffected by the hypothetical removal. The annotations confirm that this collision occurs regardless of the intervention. Solving such questions does not require reasoning about alternative outcomes or hypothetical changes. Instead, perceptual matching between the video content and the answer options is sufficient.

\subsection{Examples of the Inferential Problem}

Figure~\ref{fig:infer} illustrates a representative example of an \textit{inferential} question from the CLEVRER dataset. The question asks what event would occur if a specific object, the green cube, were removed. In this case, answering correctly requires reasoning beyond the directly observable sequence. The correct answer involves predicting a new collision that is not present in the original annotated video, namely the interaction between the yellow cylinder and the metal sphere.

This type of question cannot be resolved through direct observation alone. In the original video, the green cube initiates a chain of interactions, and its removal would alter the subsequent trajectory of the remaining objects. As such, solving the question demands counterfactual reasoning about how the physical system would evolve under a hypothetical intervention. This makes the problem fundamentally different from observational tasks and places greater demands on the model's causal understanding.

\subsection{Problem Filtering Strategy Explanation}
\label{bias_dataset}
To determine whether a visual reasoning question is \textit{observational} or \textit{interventional}, we employ a rule-based classification algorithm, detailed in Algorithm~\ref{alg:data_filtering_strategy}, to distinguish whether a given visual reasoning question is \textit{observational} or \textit{interventional}. The procedure takes as input a natural language \texttt{problem} and a candidate \texttt{option}, and proceeds in four main stages. First, the system extracts the core physical event from the option using the \texttt{ExtractBaseEvent} function. This typically corresponds to an interaction predicate such as ``the sphere and the cube collide.'' Second, the problem text is examined for linguistic negation (e.g., ``not,'' ``never'') using the \texttt{ContainsNegation} function. If negation is detected, the base event is logically inverted via the \texttt{NegateEvent} function (e.g., ``do not collide''); otherwise, the base event remains unchanged. Third, the system queries the target event—either the base event or its negation—against structured video annotations using \texttt{SearchInAnnotations}. If the event is found in the annotated data, the option is considered grounded in actual observations and the problem is classified as \textit{observational}. Otherwise, the event must be inferred under a hypothetical intervention (e.g., object removal), and the problem is labeled \textit{interventional}.

To illustrate, consider the problem: ``If the rubber object is removed, what will \textbf{not} happen?'' with the option ``The sphere and the cube collide.'' Because the question is negative, the event is negated to ``The sphere and the cube do not collide,'' and this negated event is searched for in the annotations. If it is not found, the problem requires reasoning about a counterfactual scenario and is thus classified as interventional. In contrast, for the problem ``Which of the following will happen if the cube is removed?'' and the option ``The cylinder collides with the metal sphere,'' the base event is used directly. If it is present in the annotations, the problem is classified as observational. This pipeline ensures that natural language cues, video-grounded evidence, and causal structure are systematically integrated to support robust problem categorization.

\begin{algorithm}[!t]
\caption{Data Filtering Strategy}
\label{alg:data_filtering_strategy}
\begin{algorithmic}[1]

\STATE \textbf{Input:} \texttt{problem}, \texttt{option}
\STATE \textbf{Output:} ``Observational" or ``Interference"

\STATE \COMMENT{Step 1: Extract the base event}
\STATE baseEvent $\leftarrow$ ExtractBaseEvent(option)

\STATE \COMMENT{Step 2: Determine if the problem contains negation}
\IF{problem contains negation}
    \STATE eventToQuery $\leftarrow$ NegateEvent(baseEvent)
\ELSE
    \STATE eventToQuery $\leftarrow$ baseEvent
\ENDIF

\STATE \COMMENT{Step 3: Search the event in annotations}
\STATE found $\leftarrow$ SearchInAnnotations(eventToQuery)

\STATE \COMMENT{Step 4: Classify based on search result}
\IF{found is true}
    \STATE \textbf{return} ``Observational"
\ELSE
    \STATE \textbf{return} ``Inferential"
\ENDIF

\end{algorithmic}
\end{algorithm}

\subsection{Diagnostic Experimental Setup}
We adopt the same evaluation setup as described in Section~\ref{evaluation_setting}. To eliminate the potential influence of unfamiliar answer formats on baseline model performance, we restrict our evaluation to the single-answer questions within the counterfactual subset of CLEVRER. This avoids degradation caused by exposure mismatches with multiple-answer formats. Our implementation follows Video-RFT~\cite{wang2025videorft}, using identical prompts, a sampling temperature of 0.1, top\_p of 0.001, and a batch size of 64. For each video, we sample 32 frames and upscale them to a resolution of $256 \times 28 \times 28$. The full counterfactual subset contains 9,238 questions, among which 3,945 are single-answer. To obtain fine-grained accuracy estimates, especially for inferential generalization, we evaluate model predictions at the \textit{option level} rather than only at the question level. This is necessary because some inferential questions include distractor options that are observational in nature (e.g., option B in Figure~\ref{fig:infer}). We therefore compute accuracy separately over individual options, resulting in a total of 11,524 observational options and 1,224 inferential options. This setup enables a more precise measurement of the model’s reasoning degradation and perceptual bias under fine-tuning.

\section{Discussion}
\label{limitation}
\subsection{Limitations}

Despite its promising results, our work has several limitations that offer avenues for future research. First, the effectiveness of VideoThinker is primarily validated on VQA tasks that align well with our underlying causal assumptions; its generalization to less structured tasks like video captioning remains an open question. Furthermore, our framework employs a practical, gradient-based approximation (CDPO) of a causal intervention, prioritizing computational efficiency over theoretical exactness. Finally, our analysis is centered on a 3B lightweight model where the fine-tuning degradation was most pronounced, and a more comprehensive study is needed to understand how perceptual bias and our intervention scale with larger models.

\end{document}